\documentclass[letterpaper, 10 pt, conference]{IEEEtran}
\IEEEoverridecommandlockouts
\usepackage{cite}
\usepackage{amsmath,amssymb,amsfonts}
\usepackage{algorithmic}
\usepackage{graphicx}
\usepackage{textcomp}
\usepackage{xcolor}
\usepackage{amsmath}
\usepackage{blindtext, subfig}
\usepackage{stfloats} 
\usepackage{booktabs} 
\usepackage{hyperref}
\usepackage[utf8]{inputenc}
\def\BibTeX{{\rm B\kern-.05em{\sc i\kern-.025em b}\kern-.08em
    T\kern-.1667em\lower.7ex\hbox{E}\kern-.125emX}}
\begin{document}

\title{Deep Learning Method for Cell-Wise Object Tracking, Velocity Estimation and Projection of Sensor Data over Time\\}

\author{Marco Braun\textsuperscript{1, 2}, Moritz Luszek\textsuperscript{1}, Mirko Meuter\textsuperscript{1}, Dominic Spata\textsuperscript{1, 2}, Kevin Kollek\textsuperscript{2} and Anton Kummert\textsuperscript{2}
\thanks{\textsuperscript{1}Aptiv, Am Technologiepark 1, Wuppertal, Germany}%
\thanks{\{marco.braun, moritz.luszek, mirko.meuter, dominic.spata\}@aptiv.com}%
\thanks{\textsuperscript{2}University of Wuppertal, Rainer-Gruenter-Straße 3, Wuppertal, Germany}%
\thanks{\{kollek, kummert\}@uni-wuppertal.de}%
}

\makeatletter
\def\ps@IEEEtitlepagestyle{
  \def\@oddfoot{\mycopyrightnotice}
  \def\@evenfoot{}
}
\def\mycopyrightnotice{
  {\footnotesize
  \begin{minipage}{\textwidth}
  \centering
  Copyright~\copyright~2022 IEEE.  Personal use of this material is permitted.  Permission from IEEE must be obtained for all other uses, in any current or future media, including reprinting/republishing this material for advertising or promotional purposes, creating new collective works, for resale or redistribution to servers or lists, or reuse of any copyrighted component of this work in other works. DOI: 10.1109/COINS51742.2021.9524106
  \end{minipage}
  }
}

\maketitle


\begin{abstract}
Current Deep Learning methods for environment segmentation and velocity estimation rely on Convolutional Recurrent Neural Networks to exploit spatio-temporal relationships within obtained sensor data. These approaches derive scene dynamics implicitly by correlating novel input and memorized data utilizing ConvNets. We show how ConvNets suffer from architectural restrictions for this task.\\
Based on these findings, we then provide solutions to various issues on exploiting spatio-temporal correlations in a sequence of sensor recordings by presenting a novel Recurrent Neural Network unit utilizing Transformer mechanisms. Within this unit, object encodings are tracked across consecutive frames by correlating key-query pairs derived from sensor inputs and memory states, respectively. We then use resulting tracking patterns to obtain scene dynamics and regress velocities. In a last step, the memory state of the Recurrent Neural Network is projected based on extracted velocity estimates to resolve aforementioned spatio-temporal misalignment.\\

\end{abstract}


\section{Introduction}
For automated driving functionalities, data from sensors such as camera, radar and lidar is processed to reason about environmental semantics. As each sensor scan captures a sparse subset of the environment, temporal integration of consecutive scans enriches the information density of the data and thus improves the perceptual capability of the system.\\
State of the art approaches for perceiving the environment of a vehicle \cite{long2015fully, liang_aspp, schreiber_motion_estimation, rist2020scssnet, LombacherRadar, schumann_pcs, sless_semantic_segmentation_radar, braun_quantification_of_uncertainties} utilize Deep Learning methods.
 A combination of Recurrent Neural Network (RNN) layers like Long Short-Term Memories (LSTM) \cite{hochreiter_lstm} or Gated Recurrent Units (GRU) \cite{cho_gru} and Convolutional Neural Networks (ConvNets)\cite{lecun_cnn} such as \cite{shi_conv_lstm} can be applied to integrate sensor recordings over time. These networks process a memory state, or hidden state (\textit{h}), which contains accumulated features from previous sensor scans up to \textit{t-1} as well as the novel input (\textit{I}) that contains encodings from sensor scans at time \textit{t}. By matching $h_{t-1}$ and $I_{t}$, the network is able to increase the information density of the captured environment as well as extract patterns from [$h_{t-1}$, $I_{t}$].\\ 
Scene dynamics induced by the ego-motion of the host vehicle as well as movements from external objects, however, cause spatial misalignment between $h_{t-1}$ and $I_{t}$. While knowledge about ego-motion can be used to resolve misalignment between the memory states and novel input encodings due to dynamics of the host vehicle, movements of external objects are unknown to the model and therefore harder to resolve. As a result, the model maintains spatially obsolete information in its memory which causes ambiguities and increases noise in the feature domain.
\begin{figure}
\centering
\includegraphics[width=80mm]{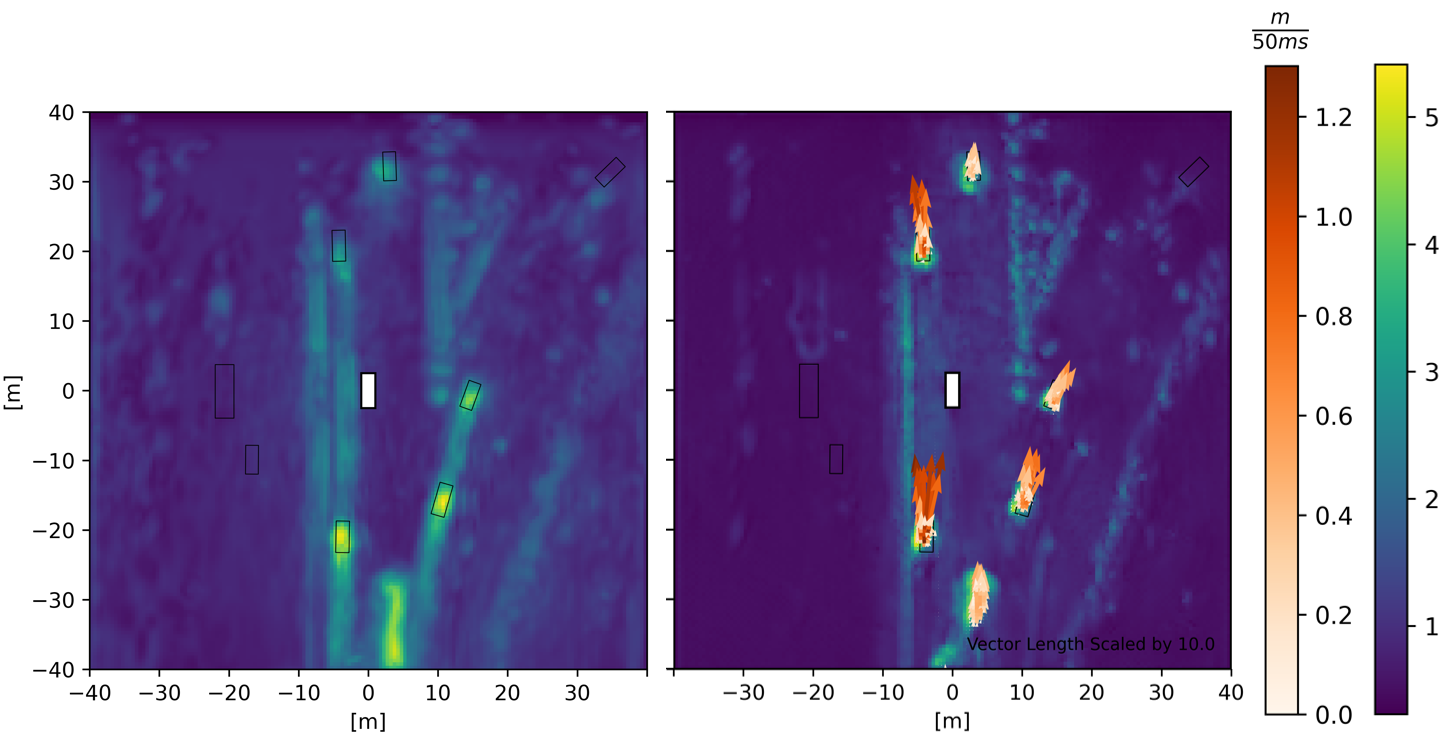}
\caption{Feature maps (L2-norm) showing the output of two Gated Recurrent Units processing sensor scans on a highway. \textit{Left}: Conventional RNN retaining trails that contain obsolete data behind moving vehicles. \textit{Right}: Our approach performs projections of the memory state based on dynamics of underlying objects to maintain spatio-temporal alignment of the data.}
\label{fig:comparison_fm}
\end{figure}
While these spatially inaccurate artifacts in the hidden state potentially reduce the prediction performance of the network, correlation between $h_{t-1}$ and $I_{t}$ may serve as a source to exploit dynamic scene context, e.g., relative movement of underlying objects between consecutive frames. Knowledge about the relative movement of underlying objects could then be utilized by the model to resolve spatio-temporal misalignment between $h_{t-1}$ and $I_{t}$. As shown in \autoref{fig:comparison_fm}, this does not happen within conventional Convolutional RNNs.\\
Another challenge when recurrently processing sensor data involves the extraction of velocities within the model. As we show in a later section of this work, the network essentially relies on spatio-temporal patterns extracted by ConvNets from [$h_{t-1}$, $I_{t}$] to predict velocities of objects. The limited receptive field size of these ConvNets thus defines the limits of object velocities that can still be perceived. This motivates the need for an advanced method to extract scene dynamics from sensor scans within the Deep Learning-based model.\\
\\
In this paper, we present a novel RNN structure for processing consecutive sensor scans that
\begin{enumerate}
\item Extracts scene dynamics from [$h_{t-1}$, $I_{t}$] on a cell basis by tracking object-related characteristics over time while being independent of the grid resolution
\item Uses these velocity patterns to resolve spatio-temporal misalignment between accumulated information from previous iterations and novel sensor scans
\end{enumerate}
This work is structured as follows: First, we summarize related work showing alternative and supplementary approaches in \autoref{sec:related_work}. We then present a novel recurrent cell that solves aforementioned issues in \autoref{sec:method}. Finally, in \autoref{sec:experiments} we show how this recurrent unit is capable of improving the performance of our network on a grid segmentation and velocity regression task.\\

\section{Related Work}\label{sec:related_work}
Our work intersects research directions of dynamic grid segmentation, optical flow for semantic video segmentation and Deep Learning-based methods for object tracking. In this section, we provide an overview about central approaches to these topics that are related to our work.\\

\subsection{Exploiting Inter-Frame Correlations}
As a central contribution of this paper, we extract scene dynamics to resolve spatio-temporal misalignments between consecutive frames. Optical flow approaches \cite{fischer_flownet, ilg_flownet_2} learn dynamics between consecutive images utilizing 2D ConvNet architectures and potentially warp these image data respectively. \cite{hur_jointopticalflow} improves temporal consistency of semantic segmentation outputs by postprocessing the network output based on optical flow estimation. However, this method processes sensor scans individually without leveraging temporal correlations within the data while we focus on assuring spatio-temporal consistency within the RNN unit.\\
In contrast, \cite{nilsson_stgru} uses a Spatial Transformer Network \cite{jaderberg_stn} to perform optical flow warping as a pre-processing step of hidden state feature maps within an RNN to effectively propagate video data over time.\\
This concept is similar to ours. However, our approach exploits scene dynamics by tracking object patterns across frames. For this purpose, we explicitly define velocities of objects within the scene that the model needs to learn as opposed to \cite{nilsson_stgru} where the network, to some precision, learns them implicitly based on the backpropagated loss signal.\\
\\
Since we present a recurrent unit that extracts movements by tracking object-related patterns between consecutive frames, our approach shows relations to tracking algorithms such as \cite{schreier2014trackingcell} that are based on Deep Learning methods like \cite{milan2016online, ningROLO, fernando2018tracking}. By combining a network like \cite{redmonyolo} to detect objects and RNN structures like LSTMs \cite{hochreiter_lstm} to exploit inter-frame correlations, these approaches discriminate targets on object level while we deploy tracking-like mechanisms to extract motion patterns from object representations within the model. To our knowledge, there is no work performing this kind of object pattern tracking on cell level to validate and refine warping of feature maps.
\subsection{Dynamic Grid Segmentation for Autonomous Driving}
Various approaches build on Deep Learning models to process lidar \cite{schreiber_motion_estimation, bieder_exploiting_multi_layer_grid_maps, schreiber_long_term_occupancy_grid_prediction, hoermann_dynamic_occupancy_grid_prediction} or radar data \cite{sless_semantic_segmentation_radar, braun_quantification_of_uncertainties} for grid segmentation, i.e. segmenting the area around the vehicle into equally sized grid cells and classifying each cell individually. While recurrently processing sensor scans allows the model to temporally integrate environmental information, RNNs that are deployed for this purpose in \cite{wu_motionnet, schreiber_motion_estimation, schreiber_long_term_occupancy_grid_prediction} show two major weaknesses: First, for moving objects, patterns from previous frames $h_{t-1}$ show a spatial misalignment to novel feature encodings $I_{t}$ from sensor scans as described above. This misalignment could be resolved by warping $h_{t-1}$ according to scene-inherent dynamics as proposed by optical flow mechanisms like \cite{nilsson_stgru}. However, this requires a reliable estimation of velocities on a cell basis.\\
Measurements from various sensors like lidar sensors do not contain intrinsic velocity information, and even radar sensors merely measure radial velocity components of detected objects. Therefore, models need to rely on movement patterns of captured objects between consecutive frames, i.e. offsets between patterns in $h_{t-1}$ and $I_{t}$ in order to extract scene dynamics. However, any ordinary Convolutional RNN like Convolutional GRUs and Convolutional LSTMs \cite{shi_conv_lstm} which process $h_{t-1}$ and $I_{t}$ has a receptive field which is restricted by the kernel size.\\
\\
\textbf{Example}: A kernel of size 3$\times$3 on a quadratic grid with a resolution of $(0.5 \hspace{1.5pt} m)^2$ and a sampling rate of 20 \textit{Hz} is able to capture movements inherent in [$h_{t-1}$, $I_{t}$] in each direction of 10 $\frac{m}{s}$. This is due to the fact that a 3x3 kernel is able to fetch information from a maximum distance of one neighboring cell in each direction towards its central grid cell within one subsequent frame. 
Even though the receptive field can be further increased by deploying bigger filters or pyramid structures as presented in \cite{wu_motionnet, schreiber_motion_estimation, schreiber_long_term_occupancy_grid_prediction}, these modifications come at a cost of increased time and space complexity while not entirely solving the problem of receptive field size dependency.\\ 
\\
In \autoref{sec:method}, we therefore present a recurrent cell that tracks objects in between frames to estimate their velocities while being independent of the underlying grid resolution. These velocities can then be used to resolve the spatio-temporal misalignment between memory state and novel sensor input.

\section{Method}\label{sec:method}

\begin{figure*}[!t]
\centering
\includegraphics[trim=5 0 10 20, clip, width=140mm]{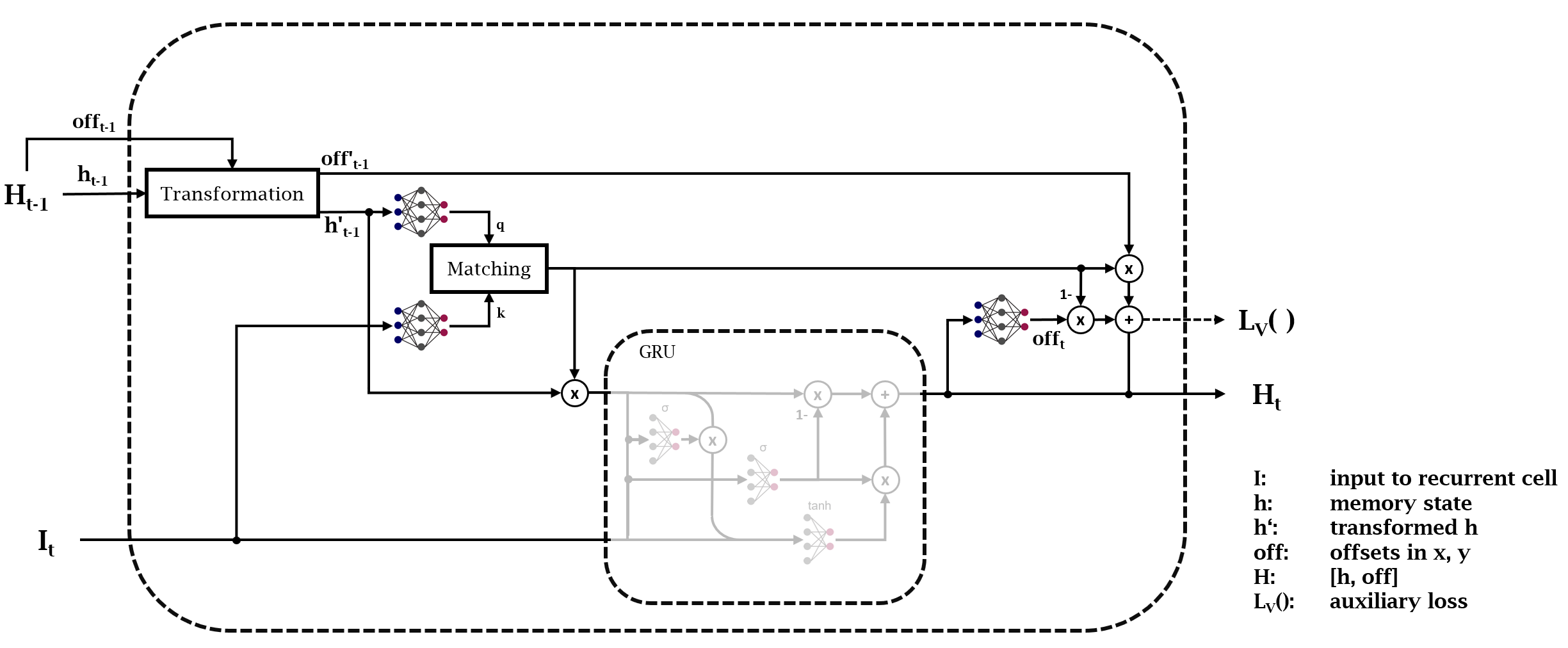}
\caption{\textbf{Recurrent State Projection Cell}: Spatio temporal misalignment caused by external scene dynamics is resolved by a transformation of $H_{t-1}$ based on regressed velocities of underlying objects $\textit{\text{off}}_{t-1}$.  This transformation is then controlled by a gating mechanism which validates the matching accuracy between embeddings of source and target patterns of the underlying objects. The resulting gated memory state is then processed with the input state $I_{t}$ by an RNN, e.g., a GRU \cite{cho_gru}. Finally, scene dynamics that determine memory state projection in the next iteration are estimated.}
\label{fig:rec_cell_diagram}
\end{figure*}

\subsection{Recurrent State Projection Cell}\label{sec:rstp} \autoref{fig:rec_cell_diagram} shows the recurrent cell that we present as a wrapper for an RNN to estimate scene dynamics and to maintain spatio-temporal alignment between the memory state $h_{t-1}$ and novel RNN inputs $I_{t}$.\\
In order to provide an initial estimate about dynamics within the scene, a velocity vector per cell is predicted by processing the hidden state $h_{t}$ using neural network layers
\begin{equation}\label{eq:speed_regression}
v_{t} = f_{\theta_{vel}}(h_{t}) 
\end{equation}
with parameters $\theta_{vel}$ that are optimized during backpropagation. For this purpose, an auxiliary regression loss is applied so that velocities can be trained in a supervised manner. Note that this initial velocity estimation will be refined recurrently to minimize the difference to actual speeds.\\
We translate predicted velocity vectors $v_{t}$ [$\frac{m}{s}$] to offset vectors $\textit{\text{off}}_{t}$ [$m$] by incorporating the frame rate of the network. For a given frame rate $FR$ [$\frac{1}{s}$], predicted offsets correspond to 
\begin{equation}\label{eq:offset_calculation}
\textit{\text{off}}_{t} = \dfrac{v_{t}}{FR}. 
\end{equation}
The predicted offset vector $\textit{\text{off}}_{t}$ is a second recurrent state alongside the hidden state $h_{t}$ so that a concatenation of $h_{t}$ and $\textit{\text{off}}_{t}$, denoted as $H_{t}$, is propagated to the next iteration.
\begin{figure}[!b]
\centering
\includegraphics[trim=340 170 280 200, clip, width=100mm]{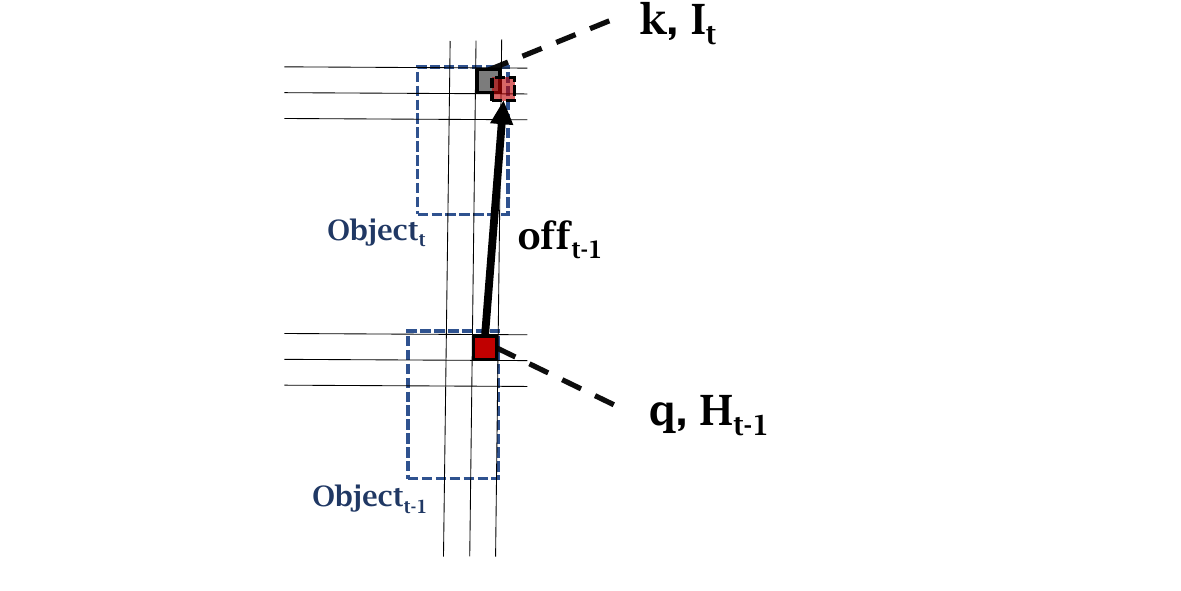}
\caption{Representation of an object which is subject to translation between two successive iterations. Object-specific encodings on a cell-basis are projected according to $\textit{\text{off}}_{t-1}$.}
\label{fig:projection_diagram}
\end{figure}
In the following iteration, the Transformation module from \autoref{fig:rec_cell_diagram} projects $H_{t-1}$ according to previously predicted projection vectors $\textit{\text{off}}_{t-1}$ (see \autoref{fig:projection_diagram}). We denote the transformed hidden state and relative offsets as $h_{t-1}'$ and $\textit{\text{off}}_{t-1}'$, respectively. For each projected data entry, we then want the model to verify the correctness of state projection to respective target locations such that correctly predicted velocities are maintained while false predictions of $v_{t}$ are suppressed. We therefore extract cell-wise embeddings as query-key pairs from both the hidden state $h_{t-1}$ and inputs to the recurrent cell $I_{t}$, where
\begin{equation}\label{eq:query}
k = f_{\theta_{q}}(I_{t})
\end{equation}
and
\begin{equation}\label{eq:key}
q = f_{\theta_{k}}(h_{t-1}).
\end{equation}
A Scaled Dot-Product Attention 
\begin{equation}\label{eq:dot_product}
Att(q', k) = \text{sigmoid}(\dfrac{q'k^{T}}{\sqrt{d_{h}}})
\end{equation}
as presented in \cite{vaswani_attention_is_all_you_need} then denotes correlations between the environment within novel input and memorized data. Here, $d_{h}$ represents the channel size of the query and key vectors per cell. Finally, before $h_{t-1}'$ is processed by a conventional GRU, it is weighted by $Att(q, k)$ so that the network is able to control information flow from source to target locations.
Finally, the second recurrent state, $\textit{\text{off}}_{t-1}'$, is recurrently refined by utilizing Att(q', k) to calculate a weighted sum 
\begin{equation}\label{eq:weighted_sum_dynamics}
\textit{\text{off}}_{t}^{+}\:=\:Att(q', k)\cdot \textit{\text{off}}_{t-1}'\:+\:(1 - Att(q', k))\cdot \textit{\text{off}}_{t}
\end{equation}
between memorized and newly predicted offsets.\\
The concept behind this mechanism can be described as follows: 
Predicting speed vectors using ConvNets within the GRU is a viable inital estimation of velocities. The matching attention $Att(q, k)$ in the following iteration correlates with the accuracy of the prediction of $\textit{\text{off}}_{t}$. Utilizing the gating mechanism from \autoref{eq:weighted_sum_dynamics}, the model is able to decide on a per cell basis to keep accurate offset predictions in memory while using newly predicted offsets in case of low correlation. This mechanism allows for a sequential refinement of predicted scene dynamics.\\
\begin{figure*}
\centering
\includegraphics[trim=150 210 180 200, clip, width=130mm]{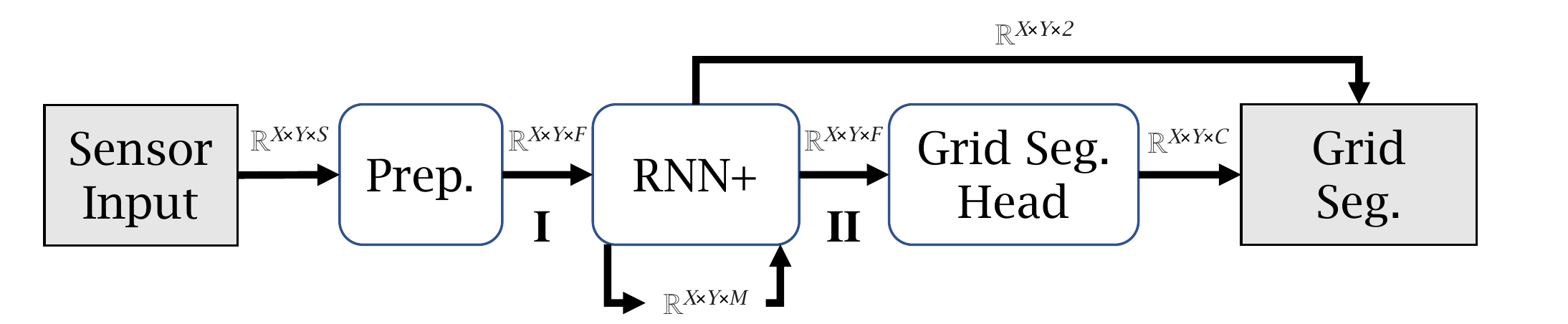}
\caption{\textbf{Network Design}: Input data in a two-dimensional Cartesian grid is processed by a preprocessing module consisting of 2D ConvNets. Resulting feature maps are then fed into a recurrent unit, RNN+, where they are processed together with 2D memory states. As depicted in \autoref{fig:rec_cell_diagram}, RNN+ outputs cell-wise velocity estimates in x and y direction together with patterns that are extracted from spatio-temporal relations in the data. Finally, a grid segmentation head consisting of Atrous Spatial Pyramid Pooling (ASPP) Layers \cite{liang_aspp} maps resulting feature maps to class probabilities for each cell.}
\label{fig:network_design}
\end{figure*}
\begin{figure}[b]
\centering
\includegraphics[trim=300 150 280 220, clip, width=90mm]{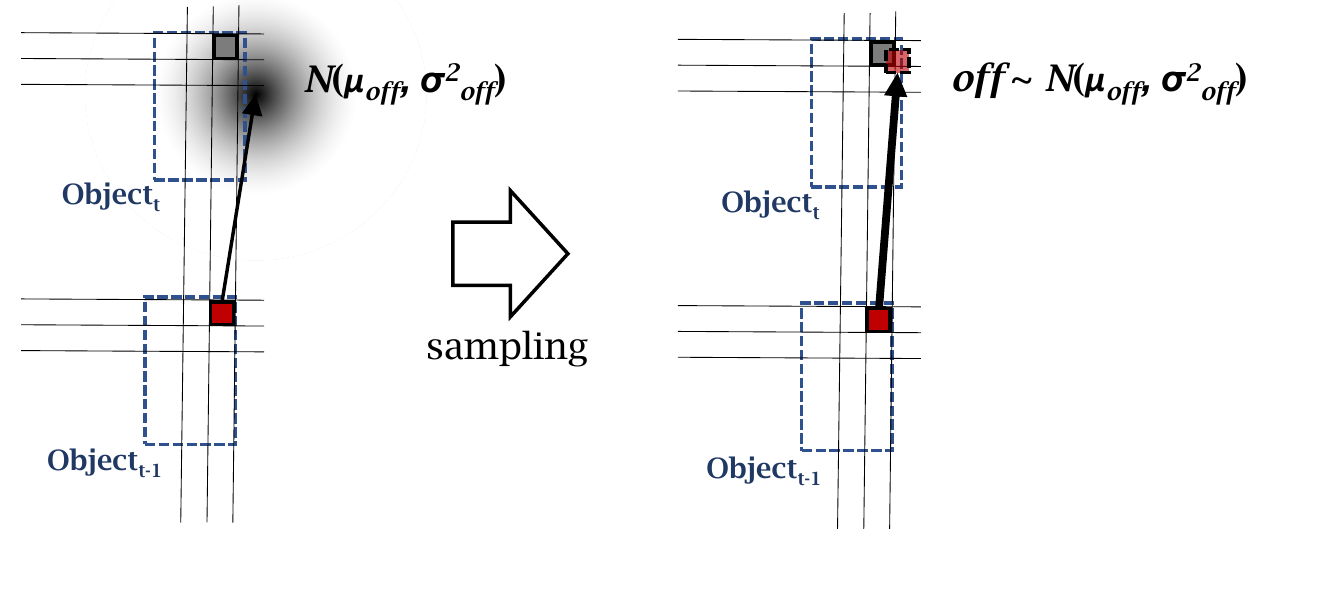}
\caption{Offsets are sampled in each iteration from joint Gaussian Distributions in X and Y.}
\label{fig:sampling_offsets}
\end{figure}
\textbf{Optional}: The presented approach can by extended by treating predicted offsets $\textit{\text{off}}$ as Gaussian distributions $\mathcal{N}(\mu_{\textit{\text{off}}}, \sigma_{\textit{\text{off}}}^{2})$ for both directions x and y where all of $\mu_{\textit{\text{off}},x}, \hspace{4pt} \sigma_{\textit{\text{off}},x}, \hspace{4pt} \mu_{\textit{\text{off}},y}$ and $\sigma_{\textit{\text{off}},y}$ are regression targets of the network output presented in \autoref{eq:speed_regression}. We assume an isotropic uncertainty distribution where $\sigma_{\textit{\text{off}},x} = \sigma_{\textit{\text{off}},y}$. Incorporating uncertainty scores into iterative refinement of scene dynamics potentially improves velocity estimation as it increases the chance of finding correlating cells for uncertain offset predictions. Loss function
\begin{equation}\label{eq:het_ale_uncertainty}
L_{V} = \dfrac{1}{N}\:\sum_{i=1}^{N}\:\dfrac{1}{2\sigma_{\textit{\text{off}}, i}^{2}}(\textit{\text{off}}_{i}'-\mu_{\textit{\text{off}}, i})^{2}\:+\:\dfrac{1}{2}\:\text{log}\:\sigma_{\textit{\text{off}}, i}^2
\end{equation}
as presented in \cite{kendall_uncertainties_in_cv} could be deployed where $N$ is the amount of training samples and $\textit{\text{off}}'$ defines the Ground Truth (GT) regression target.\\ 
Multiple offsets for each state projection could then be sampled from $\mathcal{N}(\mu_{\textit{\text{off}}}, \sigma_{\textit{\text{off}}}^{2})$ as depicted in \autoref{fig:sampling_offsets} such that multiple attention scores per emitting cell are obtained. A softmax function processing correlations for each target location then defines probabilities for various movement patterns. By projecting information based on these probabilities, the model is able to account for a variety of possible motions of external road users.


\subsection{Network Design}\label{sec:network_design}
The model we deploy to sequentially process sensor scans for dynamic grid segmentation is shown in \autoref{fig:network_design}.\\
Our approach is independent of the utilized sensor type as long as received data is available in a 2D bird's eye view grid structure. Therefore, we define an input stream of sensor data being processed by a generic Preprocessing module (Prep.). For experiments presented in this work, this Preprocessing module consists of successive 3x3 2D ConvNets. These layers transfer input sensor scans of size $\mathbb{R}^{X \times Y \times S}$ to environment encodings of size $\mathbb{R}^{X \times Y \times F}$. Here, X and Y define the amount of cells in x and y direction, respectively, and S defines the amount of sensor-specific input channels.\\
Resulting feature maps with a depth of F features are then processed by the recurrent spatio-temporal processing unit from \autoref{sec:rstp} we denote as RNN+ in \autoref{fig:network_design}. For an iteration at time \textit{t}, this module receives two data streams $I_{t}$ $\in \mathbb{R}^{X \times Y \times F}$ and $h_{t-1}$ $\in \mathbb{R}^{X \times Y \times M}$ where M denotes the amount of channels per cell in the memory state. RNN+ then outputs predictions $\textit{\text{off}}_{t}$ on cell dynamics as well as spatio-temporal patterns of size $\mathbb{R}^{X \times Y \times F}$ extracted from [$I_{t}$, $h_{t-1}$]. We then deploy a module we call grid segmentation head consisting of four successive Atrous Spatial Pyramid Pooling (ASPP) \cite{liang_aspp} layers to map outputs in II to class probabilities per cell to perform grid segmentation. These ASPP layers allow an increased receptive field for grid segmentation by using 2D ConvNets with various dilation rates in parallel.\\
\\

\section{Experiments}\label{sec:experiments}

In this section, we describe experiments we conducted to evaluate our approach as presented in \autoref{sec:method}. To this end, we describe how we preprocessed sensor data utilized for our experiments and elaborate on the training setup we applied. Finally, various mechanisms as part of the recurrent spatio-temporal processing unit defined in \autoref{sec:method} are presented and evaluated against alternative approaches building on pyramid-like network architectures to perform dynamic grid segmentation.

\subsection{Data Preprocessing}\label{data_preprocessing}

We obtain data that is used for training and evaluation of approaches presented in this section from a vehicle equipped with various sensors. We then train the model presented in \autoref{sec:method} to map input sensor scans to GT we derive from lidar recordings. In our egocentric setup, a grid of 160 x 160 cells with a cell resolution of $(0.5m)^2$ is projected around the vehicle.\\
For experiments presented in this work, we use data from a sensor returning detections on a x-y plane as input to our model.\\
Ground Truth grid semantics are derived from annotated lidar point clouds. For static environment, lidar recordings from static objects are accumulated over various past and future scans and then projected to the x-y plane. Cells are then assigned to the class that was most dominant in each grid projection. We assign the class label \textit{unknown} to any cell that is not covered by a single projection. \\
Lidar point clouds are additionally used to annotate bounding boxes for objects like vehicles. We then derive the movement of objects from the absolute shift of these bounding boxes between consecutive frames and transfer resulting velocity vectors in between object boundaries to underlying grid cells. These movement patterns are then used twofold: First, cell-wise speed vectors can directly be used for the velocity estimation. Second, cells containing objects with a velocity above 2$\frac{m}{s}$ are annotated as moving cells for grid segmentation.\\
Furthermore, we want to train our network merely on those cells that are visible to the sensors. To this end, we calculate the density of lidar rays passing each cell in the vicinity of the host vehicle. Observability maps are then created by assigning a value between 0 and 1 to each cell depending on their lidar rays coverage. These observability maps are then used as a cell-wise loss weighting during training such that unobserved cells do not contribute to the final loss.

\begin{table*}[t!]
    \small
    \centering
    \begin{tabular}{ l c c c c c c c c c c}
        \hline
        & & \multicolumn{5}{c}{Intersection over Union} & & MAE & & \\
         Recurrent Unit & & Mean & Free & Unknown & Occupied & Moving & & Velocity & & Param. \\
         \hline
         \textbf{None} & & 41.67 & 66.71 & 33.97 & 34.03 & 32.0 & & 9.29 & & 169.0K \\
         \textbf{None, aligned cap.} & & 44.21 & 70.28 & 35.85 & 36.03 & 34.7 & & 8.49 & & 369.3K \\
         \textbf{GRU} \cite{cho_gru} & & \textbf{46.61} & \textbf{74.73} & \textbf{36.62} & \textbf{37.45} & \textbf{37.65} & & \textbf{4.69} & & 359.9K \\
         \hline
    \end{tabular}
    \caption{Comparison of a model processing sensor scans individually (Single Frame) with a model exploiting temporal dimension (Multi Frame) by utilizing a GRU. Grid segmentation (IoU) and velocity estimation results (MAE) are shown.} \label{tab:temporal_integration} 
  \end{table*}

\subsection{Settings}\label{setting}
Models that are presented in this work are trained utilizing Adam Optimizer \cite{kingma_adam} with parameter settings $\beta_{1}$ = 0.9, $\beta_{2}$ = 0.999 and a learning rate of $10^{-4}$. We apply Cross Entropy loss for grid segmentation and L2 loss for cell-wise speed prediction. Due to the high imbalance between cells containing velocity values above 0 and cells with a speed regression target of 0, we deploy a cell-wise loss weighting based on global class frequency.\\
We compare models for grid segmentation by calculating the Intersection over Union (IoU) scores for each of the classes \textit{free}, \textit{occupied}, \textit{moving object} and \textit{unknown} and then average over all classes to obtain the mean IoU value (mIoU). In order to merely evaluate the models based on those cells that are observable, we only consider cells with an observability weight greater than 0 for calculating the IoU. \\
Accuracy of speed predictions is evaluated based on the Mean Absolute Error (MAE) score across both directions x and y. Here, only cells containing objects with a speed greater than 0 $\frac{m}{s}$ are considered for evaluation.\\
We train each model for 10 epochs and display results from the last epoch for comparison. \\
\subsection{Results}\label{results}

\subsubsection{Temporal Integration}\label{results_sequential_processing}
We initially stated that temporal integration of consecutive sensor scans enables the estimation of external velocities and increases the quality of network predictions due to effects like a richer coverage of the environment and the ability of the network to compensate noisy sensor measurements.\\
In order to elaborate on this assumptions, we trained three networks: The first model is comparable to the one defined in \autoref{fig:network_design} with a standard GRU \cite{cho_gru} utilized for RNN+. This network consumes 12 consecutive frames during training and is evaluated statefully.\\
We then compare this model to a second approach which processes consecutive sensor scans individually without utilizing an RNN. Since this model does not contain a memory state, it is unable to integrate information sequentially and must rely entirely on single-frame sensor scans. This model is similar to the model defined in \autoref{fig:network_design} without an RNN placed between I and II. Since this model has a reduced capacity in trainable parameters due to the missing recurrent unit, we train a third model with inflated size for better comparability. For the latter two networks, we further process feature maps in II (\autoref{fig:network_design}) by ConvNets comparable to those defined in \autoref{eq:offset_calculation} to regress velocity estimates cell-wise in x and y direction. Evaluating this additional regression target helps to understand to which extent scene dynamics can be derived from correlations between memory state and novel input within an RNN.\\
Results on these experiments, depicted in \autoref{tab:temporal_integration}, show a superior performance  of the sequentially processing approach compared to networks processing single sensor scans individually. Approaches without an RNN show a reduced IoU performance but most importantly fail to output an accurate velocity prediction. \\
These findings show the strong reliance of models on the exploitation of patterns within sequential sensor scans to capture scene dynamics.

\begin{figure*}[!b]
\centering
    \includegraphics[trim=60 60 60 60, clip, width=50mm]{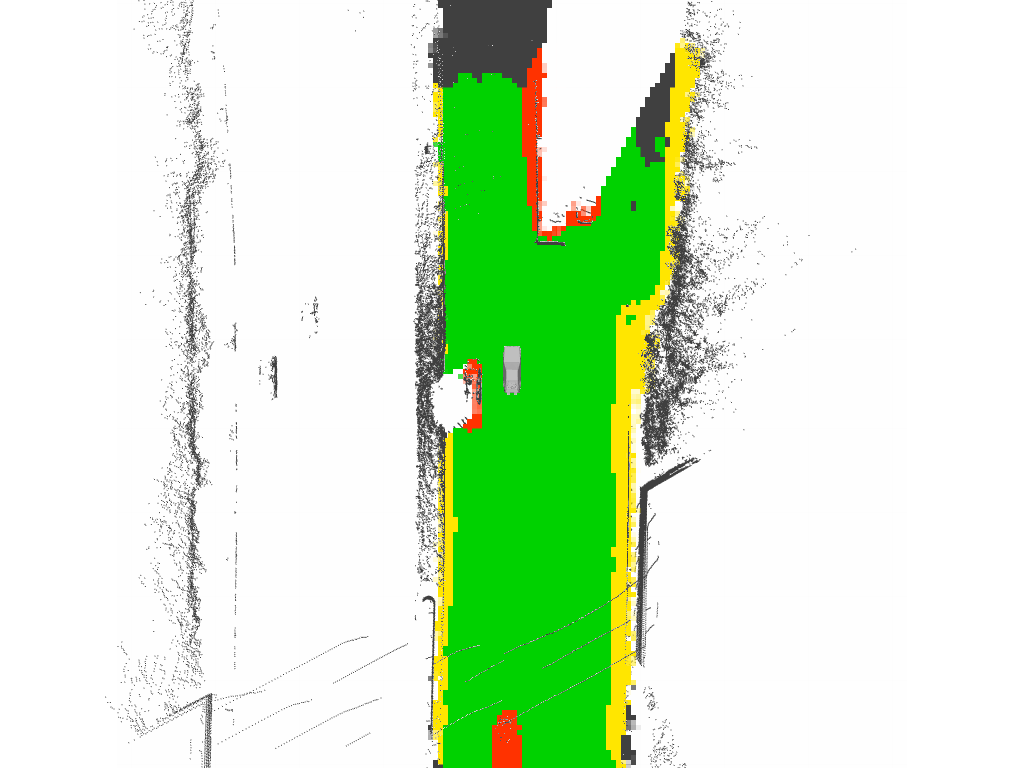}
    \includegraphics[trim=60 60 60 60, clip, width=50mm]{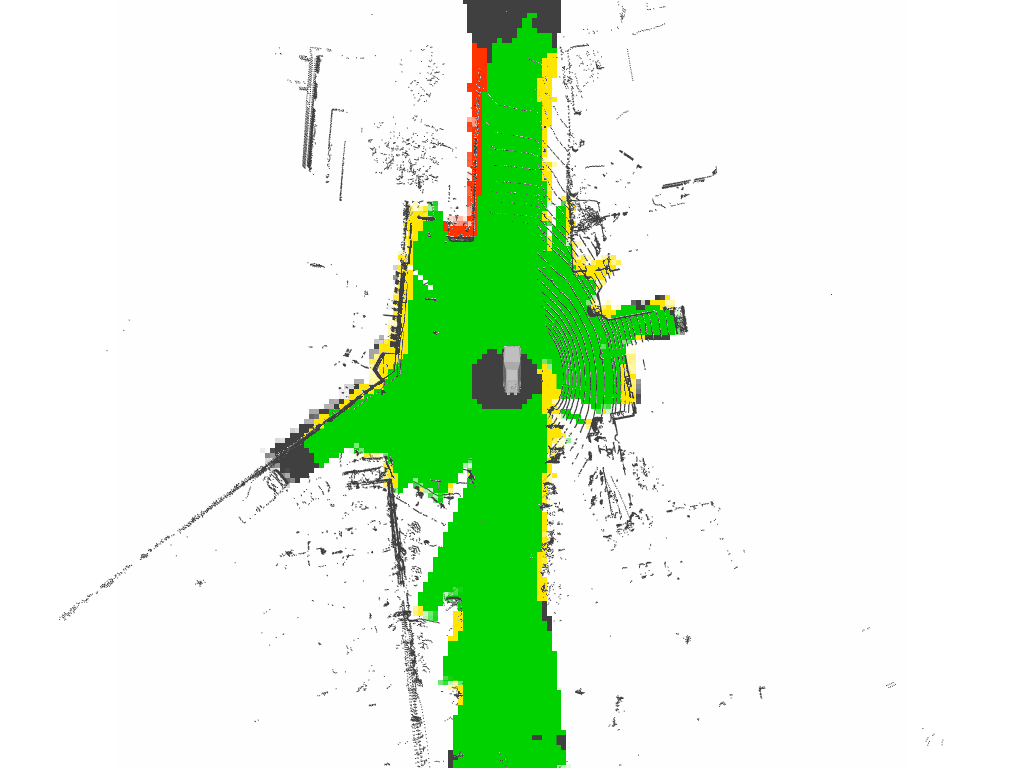}
    \includegraphics[trim=60 60 60 60, clip, width=50mm]{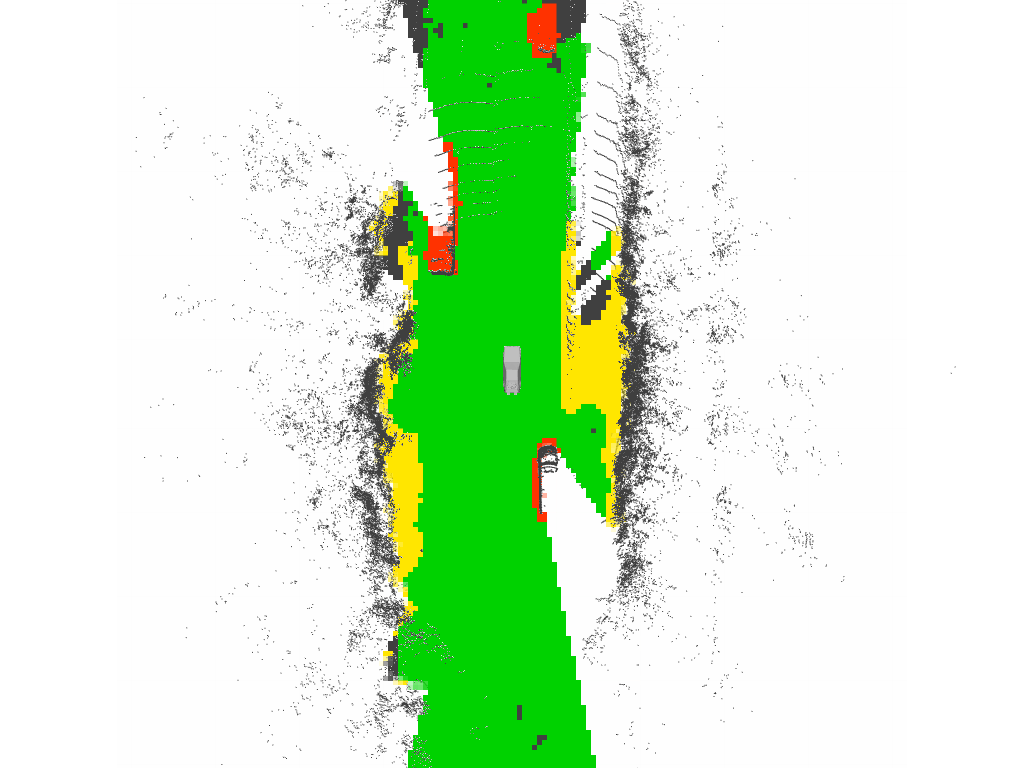}\\
    \vspace{0.1cm}
    \includegraphics[width=50mm]{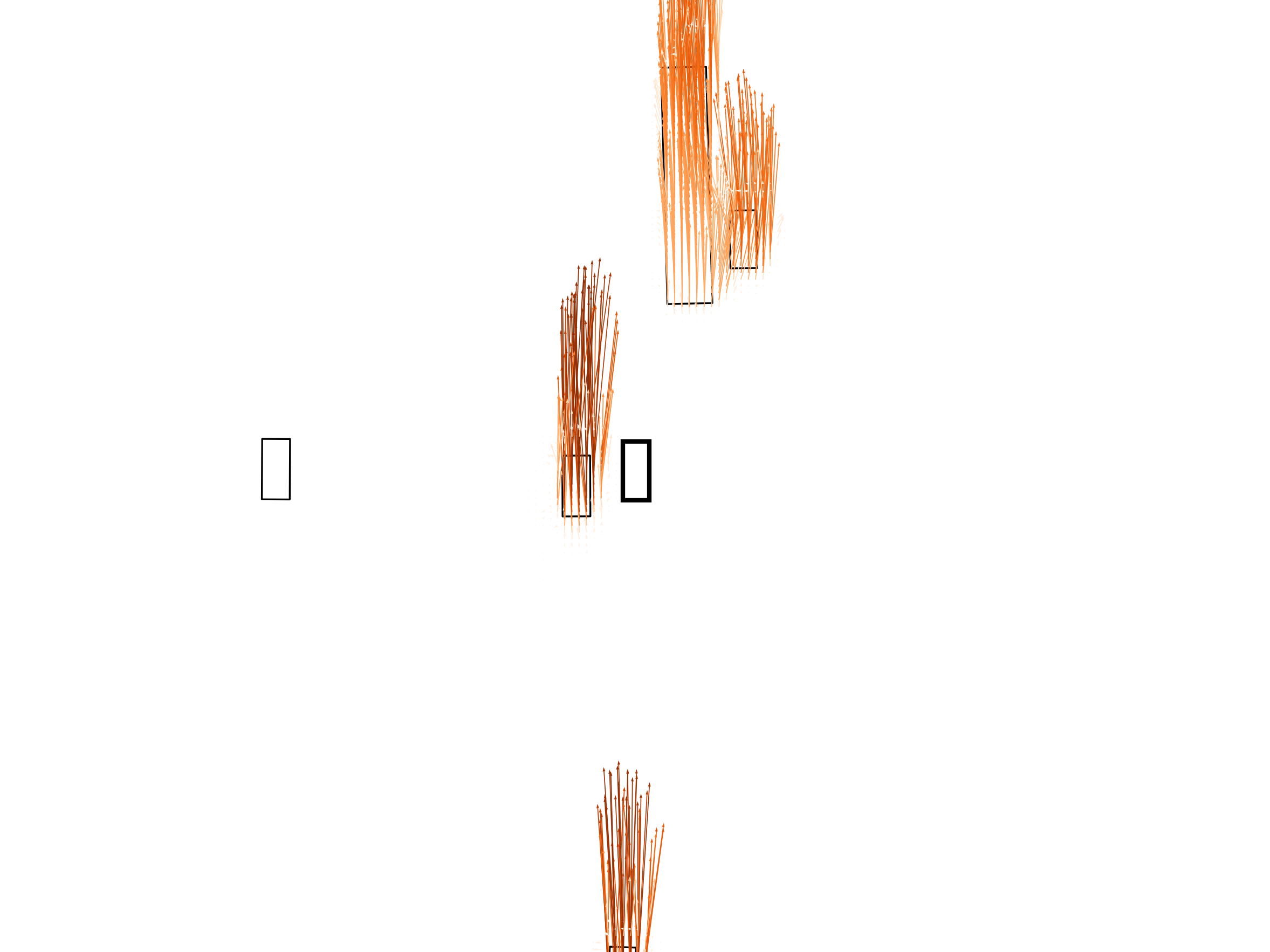}
    \includegraphics[width=50mm]{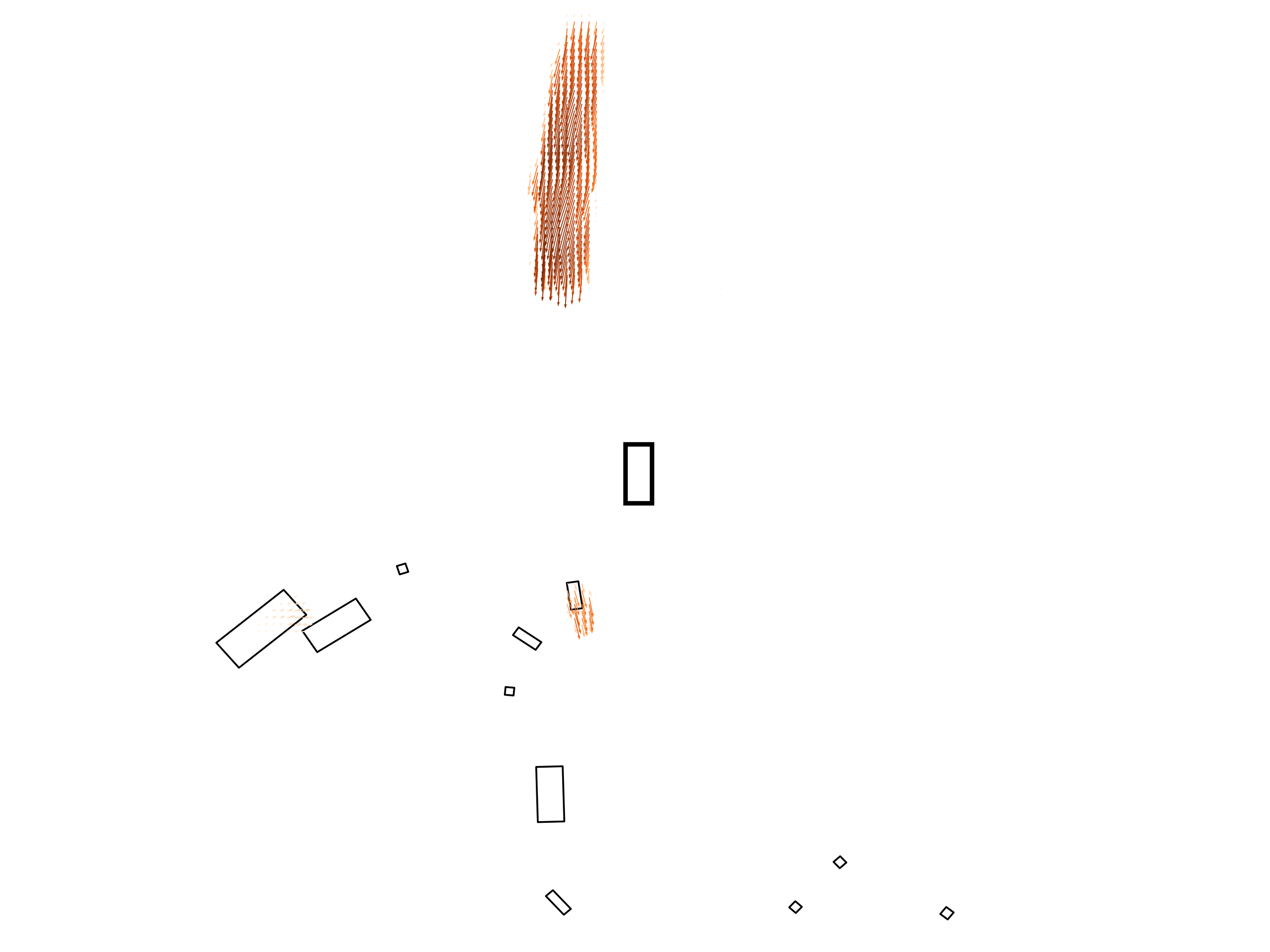}
    \includegraphics[width=50mm]{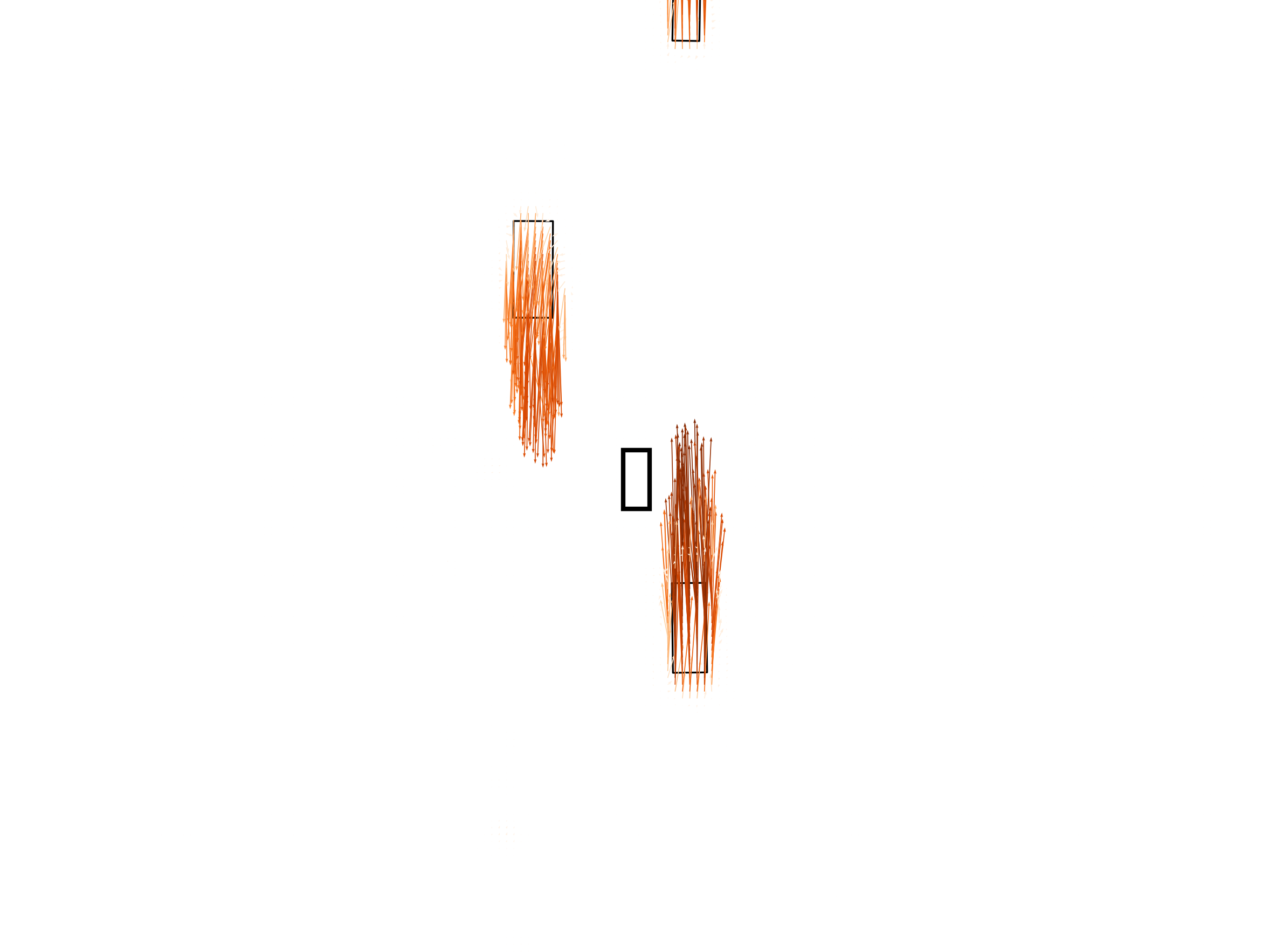}\\
\caption{Visualization of grid segmentation (top row) and cell-wise velocity estimation (bottom row) results on highway, urban and road scenes. For grid segmentation, green, yellow, grey and red cells indicate network predictions on drivable, occupied, unknown and moving areas, respectively. Cells that are not observable by the input sensor are depicted as white areas in the grid segmentation visualization.}
\label{fig:eval_graphical}
\end{figure*}

\begin{table*}[t!]
    \small
    \centering
    \begin{tabular}{ l c c c c c c c c c c}
        \hline
        & & \multicolumn{5}{c}{Intersection over Union} & & MAE & & \\
         Recurrent Unit & & Mean & Free & Unknown & Occupied & Moving & & Velocity & & Param. \\
        \hline
        \textbf{GRU}\cite{cho_gru} & & 46.61 & \textbf{74.73} & 36.62 & 37.45 & 37.65 & & 4.69 & & 359.9K \\
        \textbf{Pyramid RNN} \cite{schreiber_motion_estimation} & & 45.92 & 73.79 & \textbf{37.4} & 36.46 & 36.04 & & 4.56 & & 346.7K \\
        \textbf{Ours} & & \textbf{46.8} & 73.14 & 37.35 & \textbf{38.41} & \textbf{38.43} &  & \textbf{3.95} & & 411.6K \\
        \hline
    \end{tabular}
    \caption{Comparison of three approaches for RNN+. Grid segmentation (IoU) and velocity estimation (MAE) results are shown.} \label{tab:exploit_scene_dynamics} 
  \end{table*}

\subsubsection{Exploiting Scene Dynamics}\label{sec:exploiting_scene_dynamics}

In the previous section we showed that leveraging temporal context within the data significantly improves grid segmentation as well as scene dynamics prediction capabilities when processing sensor scans. The GRU that was deployed uses 3$\times$3 ConvNets to exploit spatio-temporal correlations in [$I_{t}$, $h_{t-1}$]. As mentioned in \autoref{sec:related_work}, these ConvNets are heavily restricted in capturing a broad range of scene dynamics due to their limited receptive field size. Approaches like \cite{schreiber_motion_estimation, wu_motionnet} partially overcome this problem by deploying ConvNets within the RNN on subsequently downsampled spatio-temporal domains such as [$I_{t}$, $h_{t-1}$]. In \autoref{tab:exploit_scene_dynamics}, we present results of such an architecture inspired by \cite{schreiber_motion_estimation} which we denote Pyramid RNN due to its pyramid-like structure. In this implementation, we use the network architecture from \autoref{fig:network_design} with RNN+ being replaced by the Multi-Scale RNN presented in \cite{schreiber_motion_estimation} with reduced amount of channels to make it comparable with the baseline model in terms of trainable parameters. \autoref{tab:exploit_scene_dynamics} shows that the larger receptive field hereby introduced leads to a small improvement in the velocity MAE compared to the baseline model utilizing a standard GRU. However, we obtain a slightly reduced grid segmentation capability.\\
We then compare the two presented approaches with the purely deterministic method defined in \autoref{sec:method}. Summing up, this method introduces the following enhancements:
\begin{enumerate}
\item Grid independent object tracking within the recurrent unit to derive scene dynamics
\item Utilize these scene dynamics to resolve the spatio-temporal misalignment between [$I_{t}$, $h_{t-1}$]
\end{enumerate}
Besides being able to exploit spatio-temporal dependencies while being independent of the grid resolution by utilizing arbitrary speed offsets, our approach explicitly formulates a tracking-based method to capture scene dynamics compared to previous methods that build on intrinsic properties of ConvNets for this task. \\
A visualization of grid segmentation as well as velocity estimation results of our method can be seen from \autoref{fig:eval_graphical}. Furthermore, \autoref{fig:comparison_fm} shows how the hidden state projection mechanism as part of our approach leads to a significant reduction of obsolete data from moving objects within the memory state of the network. \\
The quantitative improvement of our method compared to prior approaches is presented in \autoref{tab:exploit_scene_dynamics}. While this evaluation shows a slight improvement of our approach for the grid segmentation task, a superior performance with an improvement of $\sim$\textbf{ 15\% } can be obtained for velocity regression compared to approaches using an ordinary GRU or Pyramid RNN as defined in \cite{schreiber_motion_estimation}. 
In this network, for reasons of comparability, the same fundamental model as in the GRU only experiments is used while increased capacity can be attributed to the appended matching function.\\
Shown results prove that our approach combining the tracking mechanism for an enhanced velocity estimation and the state projection mechanism to resolve spatio-temporal misalignments introduced in this work present a novel and improved procedure for temporally integrating sensor data using Recurrent Neural Networks.

\section{Conclusion}

In this work, we presented a Deep Learning-based method to extract scene dynamics by tracking object-related patterns across consecutive sensor scans without being restricted to grid resolutions or velocity limits. Furthermore, we show how our method is able to utilize these velocity patterns performing a memory state projection to avoid spatio-temporal misalignment between memorized and novel data within the Recurrent Neural Network. Embeddings that are extracted by the network to validate this state projection can be interpreted as object-specific patterns and can therefore be used to track entities on a grid basis over time.\\
We show how our approach outperforms previous Recurrent Neural Network on the task of semantically segmenting the environment around a vehicle as well as velocity estimations.

\bibliography{main}{}
\bibliographystyle{unsrt}

\end{document}